%% file: main.tex
\documentclass[final,5p,times,twocolumn]{elsarticle}
\usepackage{graphics}
\usepackage{amsmath}
\usepackage{color}
\usepackage{marvosym}
\usepackage{multirow} 
\usepackage{booktabs}
\usepackage{bm}
\usepackage[switch]{lineno}
\usepackage{makecell}
\usepackage{url}
\journal{Neural Networks}
\newcolumntype{L}[1]{>{\raggedright\arraybackslash}p{#1}}
\newcolumntype{C}[1]{>{\centering\arraybackslash}p{#1}}
\newcolumntype{R}[1]{>{\raggedleft\arraybackslash}p{#1}}
\newcommand\clr[1]{{\color{black}{#1}}}

\usepackage{amsthm}

\theoremstyle{plain}

\usepackage[utf8]{inputenc} 
\usepackage[T1]{fontenc}    
\usepackage{hyperref}       
\usepackage{booktabs}       
\usepackage{amsfonts}       
\usepackage{nicefrac}       
\usepackage{microtype}      
\usepackage{xcolor}         

\usepackage{graphicx}
\usepackage{booktabs}
\usepackage{multirow}
\usepackage[normalem]{ulem}
\useunder{\uline}{\ul}{}
\usepackage{adjustbox}

\usepackage {diagbox}
\usepackage{wasysym}
\usepackage{makecell}
\usepackage{caption}

\usepackage{balance}

\usepackage{hyperref}
\usepackage{cleveref}

\usepackage{algorithm}
\usepackage{algorithmic}
\usepackage{arydshln}
\usepackage{threeparttable}
\usepackage{colortbl}

\usepackage{float}

\usepackage{listings}

\usepackage[utf8]{inputenc} 
\usepackage[T1]{fontenc}    

\begin{document}
	\begin{frontmatter}
	
\title{Self Identity Mapping}

\author[1,2]{Xiuding Cai}
\author[3]{Yaoyao Zhu}
\author[1,2]{Linjie Fu}
\author[1,2]{Dong Miao}
\author[1,2]{Yu Yao}

\affiliation[1]{organization={Chengdu Institute of Computer Application, Chinese Academy of Sciences},
	city={Chengdu},
	postcode={610213}, 
	country={China}}
\affiliation[2]{organization={University of Chinese Academic Sciences},
	city={Beijing},
	postcode={101408}, 
	country={China}}        
\affiliation[3]{organization={China Zhenhua Research Institute Co., Ltd.},
	city={Guiyang},
	postcode={550014}, 
	country={China}} 

\begin{abstract}
\clr{
	Regularization is essential in deep learning to enhance generalization and mitigate overfitting. However, conventional techniques often rely on heuristics, making them less reliable or effective across diverse settings. We propose Self Identity Mapping (SIM), a simple yet effective, data-intrinsic regularization framework that leverages an inverse mapping mechanism to enhance representation learning. By reconstructing the input from its transformed output, SIM reduces information loss during forward propagation and facilitates smoother gradient flow. To address computational inefficiencies, We instantiate SIM as $ \rho\text{SIM} $ by incorporating patch-level feature sampling and projection-based method to reconstruct latent features, effectively lowering complexity. As a model-agnostic, task-agnostic regularizer, SIM can be seamlessly integrated as a plug-and-play module, making it applicable to different network architectures and tasks.
	
	We extensively evaluate $\rho\text{SIM}$ across three tasks: image classification, few-shot prompt learning, and domain generalization. Experimental results show consistent improvements over baseline methods, highlighting $\rho\text{SIM}$'s ability to enhance representation learning across various tasks. We also demonstrate that $\rho\text{SIM}$ is orthogonal to existing regularization methods, boosting their effectiveness. Moreover, our results confirm that $\rho\text{SIM}$ effectively preserves semantic information and enhances performance in dense-to-dense tasks, such as semantic segmentation and image translation, as well as in non-visual domains including audio classification and time series anomaly detection. The code is publicly available at \url{https://github.com/XiudingCai/SIM-pytorch}.

}
\end{abstract}

\end{frontmatter}

\input{sec/1_intro}

\input{sec/2_related}

\input{sec/3_sim}
\input{sec/4_exps}
\input{sec/5_conclusion}

\section*{Code Availability}
All datasets used in this study are publicly available, and instructions for accessing them are provided in the repository. 
The code for SIM is publicly available at \url{https://github.com/XiudingCai/SIM-pytorch}.

\bibliography{main}

\input{sec/X_suppl}

\end{document}

%% file: sec/1_intro.tex
\section{Introduction}
\label{sec:intro}

Representation learning has emerged as a fundamental pillar in deep learning~\cite{bengio2013representation, ResNet, SimCLR, he2022masked, radford2021learning}, driving progress toward general artificial intelligence. At its core, representation learning aims to extract meaningful semantic from raw data, enabling models to generalize across diverse tasks. However, learning robust and transferable representations is inherently ill-posed, often requiring explicit regularization to encode domain priors and mitigate overfitting. Commonly adopted priors include smoothness~\cite{gouk2021regularisation, srivastava2014dropout}, sparsity~\cite{olshausen1996emergence, ranzato2007sparse, molchanov2017variational}, hierarchy~\cite{zhao2017pyramid, liu2021swin, cai2024rethinking}, and coherence~\cite{ho2020denoising, tulsiani2018multi}, each of which is typically enforced through dedicated regularization techniques.

Among these, weight-based regularization methods such as $\ell_1$ and $\ell_2$ penalties~\cite{goodfellow2016deep} constrain the solution space by promoting sparsity and smoothness, reducing the risk of overfitting. More dynamic approaches, such as Dropout~\cite{srivastava2014dropout}, regularize feature utilization by stochastically deactivating neurons, effectively simulating an ensemble of subnetworks and improving generalization. Extensions like Spatial Dropout~\cite{SpatialDropout} and DropConnect~\cite{wan2013regularization} refine this concept to better accommodate structured data, particularly in vision and sequential modeling tasks. However, many of these methods remain dependent on architecture~\cite{zoph2018learning, li2023dropkey} or data structure~\cite{kingma2015variational}, exhibiting varying effectiveness across different domains~\cite{kingma2015variational}. Beyond these, Batch Normalization (BN) and its variants are uniquely characterized by their reliance on statistics computed directly from the training data, such as mean and variance. This data-dependent nature enables them to adaptively regularize model activations, stabilizing training and accelerating convergence. However, different architectures~{\cite{ba2016layer, wu2018group, wu2021rethinking}} and tasks~\cite{ulyanov2016instance, huang2017arbitrary, miyato2018spectral, zhou2022generalizable} often require distinct normalization layers. Meanwhile, data augmentation~\cite{zhang2017mixup, cubuk2018autoaugment, yun2019cutmix, shorten2019survey, Gong2020KeepAugmentAS, islam2024diffusemix} enhances generalization by diversifying training samples through transformations, improving model performance on unseen data. However, these methods are often heuristically motivated and may lack generality across novel tasks, where their effectiveness and adaptability can become nontrivial or unreliable~\cite{bengio2013representation, gouk2021regularisation}.

\begin{figure}[t]
	\centering
	\includegraphics[width=0.4\textwidth]{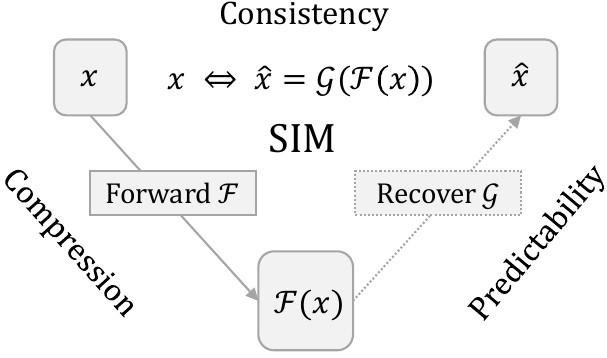}
	\caption{A simple illustration of SIM regularization. SIM achieves self-compression, self-consistency, and self-prediction through self-reconstruction.}
	\label{fig:sim}
\end{figure}

\begin{table*}[t]
	\centering
	\resizebox{0.9\textwidth}{!}{
		\begin{threeparttable}
			\centering
			\setlength{\tabcolsep}{1.2mm}{
				\begin{tabular}{lcccc}
					\toprule
					Method                          & Model-Agnostic & Task-Agnostic & Data-Intrinsic & Priors                                   \\
					\midrule
					$\ell_1$ Regularization               & \checkmark     & \checkmark    & $\times$    & Sparsity              \\
					$\ell_2$ Regularization               & \checkmark     & \checkmark    & $\times$    & Smoothness                \\
					Dropout Series~\cite{srivastava2014dropout,ghiasi2018dropblock,moon2015rnndrop,li2023dropkey}                         & $\times$     & \checkmark    & $\times$    & Smoothness                  \\
					Normalization Series~\cite{ioffe2015batch,ba2016layer, ulyanov2016instance, wu2018group}        & $\times$       & \checkmark    & \checkmark      & Smoothness \\
					Data Augmentation~\cite{zhang2017mixup,cubuk2018autoaugment}        & \checkmark     & $\times$    & \checkmark      & Smoothness       \\
					\textbf{SIM (Ours)}             & \checkmark     & \checkmark    & \checkmark    & Smoothness, Coherence, Hierarchy \\
					\bottomrule
				\end{tabular}
			}
		\end{threeparttable}
	}
	\caption{Comparison of different regularization techniques: model-agnostic, task-agnostic, and data-intrinsic properties, along with their corresponding priors.}
	\label{tab:regularizations}
\end{table*}

In response to these limitations, we propose Self Identity Mapping (SIM), a simple yet effective, data-intrinsic regularization technique. As illustrated in~\Cref{fig:sim}, SIM is designed to learn representations that are \emph{self-compressible, self-consistent, and self-predictable}. Given an input feature \( \mathbf{x}\), after undergoing a nonlinear transformation \( \mathcal{F} \), SIM aims to learn an inverse mapping \( \mathcal{G} \) such that \( \hat{\mathbf{x}} = \mathcal{G}(\mathcal{F}(\mathbf{x})) \). Modern neural networks, such as ResNet~\cite{ResNet}, are typically organized into hierarchical stages, each composed of stacked nonlinear blocks. SIM naturally integrates into these blocks, regularizing the learned features at each layer to enhance representation consistency. 

SIM offers several key advantages. First, as a data-intrinsic  regularization technique, SIM is both model-agnostic and task-agnostic and can be seamlessly integrated into diverse architectures and applying to various tasks, with minimal modifications. Second, by enforcing self-reconstruction, SIM encourages compact yet expressive representations, mitigating information loss during forward propagation. Unlike residual connections, which mitigate gradient vanishing by allowing identity mapping through shortcuts, SIM explicitly regularizes feature transformations through self-identity mapping in a complementary way. Moreover, by incorporating SIM into hierarchical neural networks, the model learns progressively refined representations, ensuring semantic consistency and preserving essential feature characteristics. As summarized in Table~\ref{tab:regularizations}, SIM provides distinct advantages over existing regularization techniques, being data-intrinsic, model-agnostic, and task-agnostic. 

While reconstruction is not a new concept and SIM shares similarities with restricted Boltzmann machines~\cite{Hinton2006AFL} and autoencoders~\cite{bengio2006greedy}, our innovation lies in extending this principle to every layer of the network as an efficient, end-to-end regularization method, rather than focusing on greedy layer-wise pretraining or feature extraction. To enhance usability, We further provide a streamlined implementation of SIM, easily adaptable to any modern networks in three simple steps (see~\Cref{sec:simple_imp}).

Nevertheless, the layer-wise reconstruction constraint in SIM incurs significant computational and memory overhead. To mitigate this, we propose a patch-level or token-level sampling strategy that selectively reconstructs local feature regions instead of the full representation, reducing computational cost. In addition, enforcing strict invertibility across all transformations can overly restrict the network's representational capacity. Motivated by recent advances in self-supervised learning~\cite{SimCLR, grill2020bootstrap, simsiam}, we introduce latent-space reconstruction, performing reconstruction in a lower-dimensional latent space via a projection layer followed by a predictor network.

By integrating the token sampling and latent-space reconstruction, we significantly enhance SIM's computational efficiency and overall performance. We refer to this approach as \( \rho\text{SIM} \). We validate \( \rho\text{SIM} \) on three tasks: image classification, few-shot prompt learning, and domain generalization, demonstrating consistent improvements across various architectures and depths. Moreover, our experiments show that \( \rho\text{SIM} \) is orthogonal to existing regularization methods, further boosting their effectiveness. We also confirm that \( \rho\text{SIM} \) effectively preserves semantic information and enhances performance in dense-to-dense tasks, such as semantic segmentation and image translation, as well as in non-visual domains including audio classification and time series anomaly detection. Finally, empirical evidence indicates that \( \rho\text{SIM} \) stabilizes the gradient norm, contributing to more robust training dynamics.

Our contributions are as follows:

\begin{itemize}
	\item \textbf{New Regularization Framework.} We present SIM, a simple yet effective data-intrinsic regularization framework that enhances feature learning through an inverse mapping mechanism. By reconstructing the input from its transformed output, SIM reduces information loss during propagation and facilitates smoother gradient flow.
	
	\item \textbf{Efficient Instantiation of SIM.} Building on this framework, we propose \( \rho\text{SIM} \), incorporating token sampling and projection-based latent feature reconstruction. These strategies substantially reduce computational complexity while preserving high performance.
	
	\item \textbf{Comprehensive Evaluation.} We extensively validate \( \rho\text{SIM} \) across a range of tasks and architectures. Results demonstrate consistent performance improvements in diverse settings, underscoring the method's versatility, robustness, and complementarity with existing regularization. 
	
\end{itemize}

The remainder of this paper is organized as follows. In Section~\ref{sec:related_work}, we review related work relevant to our study. Section~\ref{sec:sim} presents the SIM framework along with its implementation details. In Section~\ref{sec:exp}, we provide experimental results demonstrating the effectiveness and generality of SIM across various tasks and architectures. Finally, Section~\ref{sec:conclusion} concludes the paper and discusses potential future directions.

%% file: sec/2_related.tex
\section{Related Works}
\label{sec:related_work}

\begin{figure*}[t]
	\centering
	\includegraphics[width=1.\textwidth]{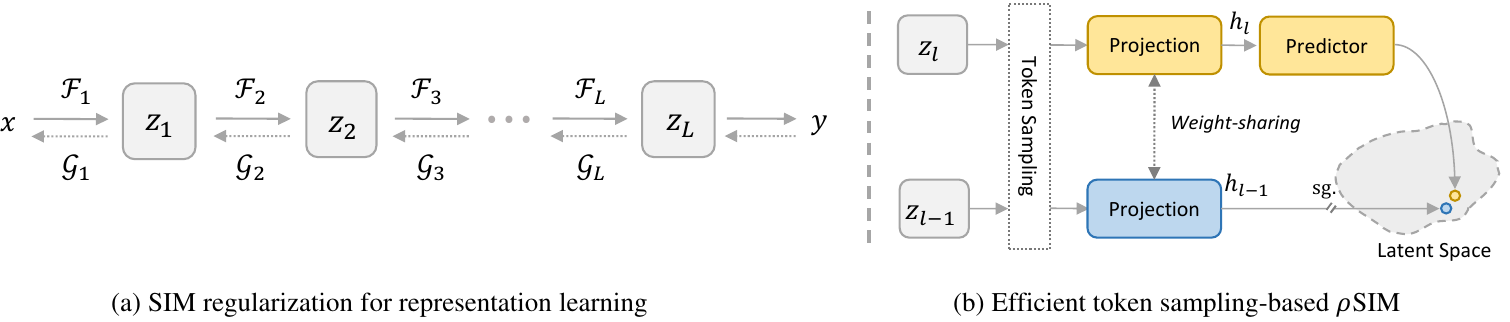}
	\caption{Overall framework of SIM.}
	\label{fig:arch}
\end{figure*}

\subsection{Regularization}

Regularization is a fundamental technique in deep learning that aims to enhance model generalization and prevent overfitting~\cite{bengio2013representation, moradi2020survey}. Existing regularization methods can be broadly categorized into three types: parameter regularization, structural regularization, and data augmentation.

Parameter regularization techniques impose constraints on model parameters to control complexity~\cite{goodfellow2016deep}. $\ell_1$ regularization penalizes the absolute value of weights, promoting sparsity and aiding feature selection. $\ell_2$ regularization applies a squared penalty, encouraging smaller weights to mitigate overfitting. Weight decay~\cite{hanson1988comparing}, a variant of $\ell_2$, further constrains weight growth, leading to improved generalization, particularly in deep networks. Despite their effectiveness, these methods often fall short in capturing the intricate, high-dimensional dependencies within modern architectures.

Structural regularization methods improve model generalization by modifying the network architecture. Dropout~\cite{srivastava2014dropout} is a classic technique that randomly deactivates neurons during training, reducing reliance on specific units and enhancing robustness. Initially applied to fully connected layers, Dropout has been adapted for various architectures, including Convolutional Neural Networks~\cite{SpatialDropout, ghiasi2018dropblock, pham2021autodropout}, Recurrent Neural Networks~\cite{moon2015rnndrop, semeniuta2016recurrent, molchanov2017variational}, and Transformers~\cite{fan2019reducing, li2023dropkey}. Batch Normalization~\cite{ioffe2015batch} standardizes each mini-batch, mitigating internal covariate shift and providing a regularizing effect. Batch Normalization~\cite{ioffe2015batch} and its variants~\cite{ba2016layer, ulyanov2016instance, wu2018group} standardize activations across different dimensions, mitigating internal covariate shift and introducing a regularizing effect. Although effective in practice, the performance of these methods is often heuristic, making them highly sensitive to hyperparameter tuning and tailored to specific tasks or architectures~\cite{bengio2013representation, gouk2021regularisation}.

Data augmentation enhances model generalization by diversifying training data. Traditional methods, such as rotation, scaling, and cropping, simulate common input variations. Recent approaches, like Mixup~\cite{zhang2017mixup} and CutMix~\cite{yun2019cutmix}, generate new examples by combining multiple samples, promoting smoother decision boundaries and improving robustness. Automatic strategies, such as AutoAugment~\cite{cubuk2018autoaugment} and RandAugment~\cite{cubuk2020randaugment}, leverage search algorithms to optimize augmentation policies, further boosting generalization. However, these techniques often demand extensive tuning or computational resources, and their performance may vary with task or dataset.

\subsection{Reconstruction}

Reconstruction plays an important role in representation learning by aiming to map input data into a latent space and then reconstruct the original input, thereby improving feature representation quality~\cite{he2022masked, gui2024survey, lu2024anomaly}. 

The Autoencoder~\cite{Hinton1993AutoencodersMD, zhai2018autoencoder} is a foundational method in this domain, consisting of an encoder-decoder architecture to minimize the discrepancy between input and reconstruction. Variants like Variational Autoencoder~\cite{kingma2013auto} use probabilistic modeling for both reconstruction and sample generation, while Denoising Autoencoder~\cite{Vincent2008ExtractingAC} enhances robustness by reconstructing clean data from noisy inputs. The Sparse Autoencoder~\cite{ranzato2007sparse} enforces sparsity to learn high-dimensional representations. Stacked Autoencoders (SAE) train autoencoders layer by layer to refine hierarchical features~\cite{bengio2007greedy}. This unsupervised pretraining improves initialization for downstream tasks but primarily serves feature extraction rather than direct regularization. 

Beyond autoencoders, reconstruction is central to self-supervised learning, where models leverage input reconstruction as a pretext task to capture intricate structures and learn more expressive, transferable features, ultimately improving generalization across diverse tasks~\cite{devlin2019bert, he2022masked, wang2023frequency, wang2025non}.

Reconstruction is also utilized for task-specific regularization. For instance, CycleGAN~\cite{CycleGAN2017} introduced cycle consistency, where generated images are mapped back to their original domain to preserve semantic structure. Similarly, Gaurav Parmar et al.~\cite{parmar2023zero} introduced cross-attention map consistency for prompt-based image translation tasks to ensure structural consistency in generated images. {EnCo}~\cite{cai2024rethinking} employs encoder-decoder symmetry, reconstructing encoder features from decoder representations to maintain content constraints. Deep Image Prior (DIP)~\cite{ulyanov2018deep} leverages the convolutional network structure for image reconstruction, achieving tasks such as denoising and super-resolution without extra training data. By minimizing reconstruction loss, it delivers high-quality results with minimal supervision.

In contrast to  traditional reconstruction methods, SIM operates without altering network architectures or relying on specific data types or predefined tasks. Instead, SIM enforces input reconstruction at the block level via a self-identity mapping mechanism. As a plug-and-play module, SIM integrates seamlessly into any network, enhancing representation learning and improving generalization.

%% file: sec/3_sim.tex
\section{Self Identity Mapping}
\label{sec:sim}

\subsection{Overview of SIM}

In this work, we introduce SIM, a simple yet effective approach designed to preserve input information after each nonlinear transformation block in a neural network. The concept is straightforward: for an input $\mathbf{x}$ passing through a nonlinear block $\mathcal{F}$, we obtain $\mathbf{z} = \mathcal{F}(\mathbf{x})$. Our objective is to recover $\mathbf{x}$ by applying another nonlinear transformation, $\hat{\mathbf{x}} = \mathcal{G}(\mathcal{F}(\mathbf{x}))$. This can be viewed as a self-supervised mechanism that ensures the network retains essential input information at each layer.

Formally, we view $\mathcal{F}$ and $\mathcal{G}$ as an encoder-decoder pair, where $\mathbf{z}$ represents the latent feature after encoding. Consider a neural network with $L$ nonlinear blocks (as shown in Figure~\ref{fig:arch} (a)), where the output of the $l$-th block is denoted as $\mathbf{z}_{l} = \mathcal{F}_{l}(\mathbf{z}_{l-1})$, with $\mathbf{z}_{l-1}$ being the input to the $l$-th layer. Notably, each \(\mathcal{F}_{l}\) can be regarded as the encoder component of an autoencoder, so we introduce the corresponding decoder \(\mathcal{G}_{l}\) to reconstruct the input, i.e. \(\hat{\mathbf{z}}_{l-1}=\mathcal{G}_{l}(\mathbf{z}_{l})\). In this manner, SIM establishes a self-identity mapping mechanism at every layer, ensuring that essential input information is consistently preserved. To enforce and quantify this consistency across layers, we use the mean squared error (MSE) to measure the differences as follows:
\begin{equation}
	\mathcal{L}_{\text{SIM}} = \sum_{l=1}^{L} \text{MSE}\Big(\mathtt{stopgrad}(\mathbf{z}_{l-1}), \mathcal{G}_l(\mathbf{z}_l)\Big),
\end{equation}
where $\mathtt{stopgrad}$ is applied to block gradient flow from earlier layers, preventing the leakage of supervisory signals.

\subsection{Efficient SIM with Token Sampling and Projection}

SIM enables hierarchical, progressively refined feature representations by encouraging self-reconstruction at each block. However, the inclusion of the auxiliary decoder \(\mathcal{G}\) significantly increases model complexity and computational cost. To address this, we propose an efficient version of SIM utilizing token sampling and projection layers, as illustrated in Figure~\ref{fig:arch} (b). 

For any 2D input $\mathbf{z}_{l} \in \mathcal{R}^{B \times C\times H \times W}$ at layer $l$, we first reshape it to $\mathbf{z}_{l}^{\prime} \in \mathcal{R}^{B \times N \times C}$, where $N = H \times W$. To exploit the locality of the input, we sample a fraction \(\rho\) of tokens from the feature map for reconstruction, yielding \(\dot{\mathbf{z}}_{l} \in \mathcal{R}^{B \times n \times C}\), where \(n = \rho N\) and \(\rho \in (0, 1]\). This method, referred to as \(\rho\)SIM, reduces computational complexity while maintaining essential reconstruction.

\begin{algorithm}[t]
	\caption{Efficient Implementation of $\rho$SIM}
	\label{alg:rhoSIM}
	\definecolor{codeblue}{rgb}{0.25,0.5,0.5}
	\definecolor{codekw}{rgb}{0.85, 0.18, 0.50}
	\lstset{
		backgroundcolor=\color{white},
		basicstyle=\fontsize{9pt}{9pt}\ttfamily\selectfont,
		columns=fullflexible,
		breaklines=true,
		captionpos=b,
		commentstyle=\fontsize{9pt}{9pt}\color{codeblue},
		keywordstyle=\fontsize{9pt}{9pt}\color{codekw},
	}
	\begin{lstlisting}[language=python]
from SIM import init_SIM, forward_with_SIM
	
class ResNetBlock(nn.Module):
	
    @init_SIM()  # step 1
    def __init__(self, in_channels, out_channels):
        # pass
    
    @forward_with_SIM()  # step 2
    def forward(self, x):
        # pass
    
# Model training with rho_SIM
model = ResNet50()
optimizer = torch.optim.Adam(model.parameters())
criterion = nn.CrossEntropyLoss()
    
for data, label in dataloader:
    optimizer.zero_grad()
    output = model(data)
    loss = criterion(output, label)    
    sim_loss = model.sim_loss  # step 3
    total_loss = loss + lambda_sim * sim_loss        
    total_loss.backward()
    optimizer.step()
    \end{lstlisting}
\end{algorithm}

Inspired by recent self-supervised progress~\cite{SimCLR, grill2020bootstrap, simsiam}, we further introduce two projection layers, $\mathcal{H}_1(\cdot)$ and $\mathcal{H}_2(\cdot)$, to enhance feature representation. These layers project the sampled features \(\dot{\mathbf{z}}_{l-1}\) and \(\dot{\mathbf{z}}_l\) into a $D$-dimensional space, defined as:
\begin{equation}
	{\mathbf{h}}_{l-1} = \mathcal{H}_1(\dot{\mathbf{z}}_{l-1}), \quad \mathbf{h}_l = \mathcal{H}_2(\dot{\mathbf{z}}_l),
\end{equation}
where \(\mathbf{h}_{l-1}\) and \(\mathbf{h}_l \in \mathcal{R}^{B \times n \times D}\). To stabilize training, we normalize the token-level features by computing the mean and variance across the \(n\) tokens, resulting in the normalized features,
\begin{equation}
	\tilde{\mathbf{h}}_l = \frac{\mathbf{h}_l - \mu(\mathbf{h}_l)}{{\sigma(\mathbf{h}_l) + \epsilon}},
\end{equation}
where \(\mu(\mathbf{h}_l)\) and \(\sigma(\mathbf{h}_l)\) are the mean and standard deviation computed over the \(n\) tokens, and \(\epsilon\) ensures numerical stability. The final loss function for the $\rho$SIM becomes:
\begin{equation}
	\mathcal{L}_{\text{SIM}} = \sum_{l=1}^{L} \text{MSE}\Big(\mathtt{stopgrad}(\tilde{{\mathbf{h}}}_{l-1}), \mathcal{G}_l(\tilde{{\mathbf{h}}}_{l})\Big).
\end{equation}

\subsection{Simple Implementation} 
\label{sec:simple_imp}
SIM regularizes features through simple self-reconstruction. In the implementation of $\rho$SIM, the projection is realized using a straightforward three-layer fully connected network with ReLU activations, while the predictor is a single linear layer. We find that this simple setup allows $\rho$SIM to perform effectively.

Modern neural networks are typically composed of homogeneous or heterogeneous nonlinear blocks. To further simplify the application of $\rho$SIM, we implemented it using a decorator-based approach. This enables the use of SIM in just three steps: adding decorators to the initialization and forward propagation functions of the target blocks, and incorporating the SIM loss into the optimization process. A Torch-like example is provided in \Cref{alg:rhoSIM}.

%% file: sec/4_exps.tex
\section{Experiments}
\label{sec:exp}

\subsection{Image Classification}

To evaluate the efficiency and effectiveness of the SIM, we conducted extensive experiments on four widely-used benchmark datasets: CIFAR10, CIFAR100, SVHN, and CUB. In these experiments, our primary objective is to systematically assess the impact of SIM on established network architectures, rather than to push new state-of-the-art (SOTA) performance. 

\input{tables/exp_cifar_ext}

\paragraph{Implementation}  For all backbone networks, we augment their blocks with corresponding $\rho$SIM mechanisms. Specifically, for every block (\emph{e.g.}, a ResNet block), we introduce an auxiliary SIM loss that encourages feature extraction to retain and compress essential information before passing it to the subsequent block. The total SIM loss is computed as the sum of all block-wise SIM losses and incorporated into the overall training objective, with a token sampling rate $\rho$ of 0.2 and a SIM loss weight $\lambda$ of \(5 \times 10^{-3}\).

\paragraph{CIFAR10, CIFAR100 and SVHN Results}  CIFAR10 and CIFAR100 consist of 50,000 training images and 10,000 testing images, with 10 and 100 classes, respectively. The SVHN dataset is designed for real-world digit recognition, containing over 600,000 images of house numbers extracted from Google Street View. See Appendix~\hyperref[sec:appendix_D]{D} and \hyperref[sec:appendix_E]{E} for dataset details and training settings. Table~\ref{tab:cifar} summarizes the performance of ResNets of varying depths (ranging from 18 to 152 layers) under different regularization strategies, with and without the incorporation of SIM. All experiments were conducted with three different random seeds, and the reported results represent the average performance.

Our results reveal several key insights. First, incorporating SIM consistently enhances performance across all tested ResNet depths, indicating that the method robustly improves feature representation and alleviates overfitting. This trend is evident on CIFAR10, CIFAR100, and SVHN, suggesting that SIM effectively adapts to diverse data characteristics. Second, our analysis shows that SIM is orthogonal to common regularization techniques such as Mixup, CutMix, and Label Smoothing. This orthogonality is crucial, as it demonstrates that SIM can be seamlessly integrated with these techniques, yielding additive benefits. For instance, on the CIFAR100 dataset, combining SIM with CutMix resulted in an absolute improvement of nearly 0.57 percentage points compared to using CutMix alone. Moreover, we also report the parameter counts for networks of different depths revealing that SIM adds only a negligible number of extra parameters relative to the total model size.

\input{tables/exp_cub}

\input{tables/exp_few_shot}

\input{tables/exp_dg_and_beta}

\paragraph{CUB Results}  The CUB dataset is a fine-grained visual classification benchmark, featuring 200 bird species, with approximately 100 images per class and high-quality annotations. As a challenging fine-grained dataset, CUB requires models to capture subtle inter-class variations and visually discriminative features. 

In our experiments, we evaluated the impact of SIM across a range of network architectures, including ResNet~\cite{ResNet} and DenseNet~\cite{huang2017densely}, the modern convolutional network ConNeXt~\cite{liu2022convnet}, and the Swin Transformer~\cite{liu2021swin}. As shown in~\Cref{tab:cub}, integrating SIM consistently enhanced accuracy across these diverse models. Notably, for ConNeXt, the incorporation of SIM increased accuracy from 89.68\% to 91.80\%, which represents an absolute gain of 2.21\% and demonstrates its potential to significantly improve performance in challenging classification tasks. 

We also conducted a detailed analysis of computational efficiency by reporting both the parameter count and FLOPs (in billions). Consistent with our observations on the CIFAR datasets, SIM introduces only a minimal computational overhead. For instance, when applied to ConNeXt, SIM increased the required FLOPs by a factor of only 1.048, ensuring improved generalization without compromising computational feasibility.

\subsection{Few-Shot Prompt Learning}

In the previous experiments, we demonstrated that incorporating SIM into existing network blocks effectively mitigates overfitting and enhances model performance. In this section, we investigate the effectiveness of SIM in the efficient fine-tuning of large-scale models, particularly in the context of few-shot prompt learning. Few-shot learning requires a model to adapt using only a limited number of labeled samples, making it crucial to extract essential feature representations while suppressing noise. This setting presents a more stringent test for evaluating SIM's capacity to enhance generalization under data-scarce conditions.

For our experiments, we use the CLIP (Contrastive Language-Image Pre-training) model~\cite{radford2021learning} as the backbone. CLIP, an unsupervised vision-language model, has demonstrated remarkable performance in a variety of downstream tasks. However, despite its impressive zero-shot capabilities, CLIP still struggles in few-shot settings when only a limited number of labeled samples are available. Recent SOTA methods have shown that fine-tuning a small number of parameters, or leveraging a few samples, can significantly enhance CLIP's performance in few-shot scenarios.

\paragraph{Baseline Methods}  We selected two commonly used few-shot prompt learning baseline methods for comparison: Tip-Adapter					~\cite{zhang2021tip} and Meta-Adapter~\cite{song2023meta}. Both methods use lightweight adapter modules to fine-tune pre-trained models for few-shot tasks. Tip-Adapter adapts the model with simple modules, while Meta-Adapter optimizes feature representations with residual adapters to reduce overfitting and improve inference speed. SIM integrates seamlessly with these adapter-based methods by applying the self-identity mapping mechanism to each adapter, further improving their generalization ability when faced with limited samples.

\paragraph{Comparison Results}  We report the performance of various few-shot prompt learning methods across seven commonly used datasets in~\Cref{tab:few_shot}. Results show that SIM consistently improves Tip-Adapter's performance across all datasets, increasing the average accuracy from 73.83\% to 74.21\%. For Meta-Adapter, SIM boosts performance on most datasets, except for a slight drop on the OxfordPets dataset, achieving an average accuracy increase from 56.14\% to 57.06\%. These results demonstrate that SIM consistently enhances fine-tuning performance under few-shot conditions. We attribute this to SIM's role as a task-agnostic regularization term, which helps extract more robust features. Furthermore, SIM's reconstruction mechanism provides smoother gradient flows, facilitating model optimization. For further discussion on gradient behavior, refer to~\Cref{sec:grad_norm}.

\subsection{Domain Generalization}

Domain Generalization (DG) focuses on training models that generalize well to unseen target domains without additional domain-specific training. In this section, we examine the performance of SIM on domain generalization tasks and validate its effectiveness in enhancing the model's cross-domain generalization ability.

\input{tables/exp_i2i}

\input{tables/exp_drive}

\paragraph{Datasets and Evaluation Metrics}  We used two commonly used domain generalization datasets: PACS~\cite{li2017deeper} and TerraIncognita~\cite{beery2018recognition}. The PACS dataset contains four source domains (art paintings, photos, cartoons, and sketches) with corresponding multi-class labels to test the model's ability to generalize across domains. The TerraIncognita dataset includes five source domains with large visual style differences between domains. To evaluate the performance of different methods in domain generalization, we used accuracy (ACC) as the evaluation metric to measure classification performance across multiple target domains.

\paragraph{Baseline Methods}  To thoroughly evaluate SIM in domain generalization tasks, we selected several representative baseline methods, including IRM~\cite{IRM}, MixStyle~\cite{mixstyle}, and VREx~\cite{VREx}. We also chose ERM, which directly trains the model on source domains and ignores target domain differences. This serves as the simplest baseline method for comparison with more complex methods.

\paragraph{Comparison Results}  As shown in~\Cref{tab:dg}, experiments on the PACS and TerraIncognita datasets demonstrate that SIM consistently outperforms existing baseline methods in domain generalization tasks. Notably, on the PACS dataset, the combination of ERM and SIM surpasses the performance of all other DG methods. Similarly, on the TerraIncognita dataset, ERM+$\rho$SIM outperforms MixStyle, a method specifically designed for domain generalization. These results underscore SIM's effectiveness as a task-agnostic regularization method, capable of enhancing feature representation and preserving core feature consistency, thereby improving the model's generalization across domains.

\subsection{Ablations on $\rho$SIM}

\paragraph{Token Sampling Ratio} 
We first investigate how the token sampling ratio $\rho$ affects SIM by using $\rho=0$ (no SIM) as a baseline in~\Cref{fig:ablation} (left). The results show that moderate values of $\rho$, particularly around 0.2 and 0.6, consistently yield higher accuracies on both CIFAR-10 and CIFAR-100. However, once $\rho$ exceeds 0.7, performance degrades noticeably, likely because an excessive sampling rate restricts the model's ability. In practice, we choose $\rho=0.2$ by default for its efficiency.

\paragraph{SIM Loss Weight} 
Next, we investigate the impact of the SIM loss weight $\lambda$ by varying it from $10^{-5}$ to 1, as shown in Figure~\ref{fig:ablation} (right). A value of $\lambda = 0$ corresponds to the baseline without SIM. We observe that an appropriate range for $\lambda$ (from $10^{-5}$ to $10^{-2}$) yields performance gains. However, when $\lambda$ becomes too large, the constraint becomes overly restrictive, limiting the model's expressive power. For our experiments, we set $\lambda$ to 0.005 by default.
 
\subsection{Can SIM Help Semantic Retention?}

In this subsection, We investigate whether SIM regularization facilitates semantic retention by evaluating its impact on two dense-to-dense tasks: image-to-image translation and image segmentation. Extensive experiments across diverse architectures and datasets demonstrate that SIM leads to more consistent and robust semantic representations.

\paragraph{Impact on Image-to-Image Translation} To assess SIM's effectiveness in preserving structural details, we begin by examining its role in image-to-image translation tasks. Using CUT~\cite{park2020cut} as our baseline, where content consistency is traditionally enforced via an NCE loss, we substitute this loss with SIM regularization in the generator network. This modification yields substantial improvements: on the Cat$\rightarrow$Dog dataset~\cite{choi2020stargan}, FID decreases from 76.2 to 58.2 (23.6\% relative improvement), and on Cityscapes~\cite{cordts2016cityscapes}, mAP increases from 24.7\% to 28.1\%. Notably, GANs training typically collapse, as shown in~\Cref{fig:pca}~(a). However, SIM regularization stabilizes training and preserves structural semantic. PCA visualizations in Figure~\ref{fig:pca} further illustrate that SIM produces more coherent and semantically consistent feature representations.

\paragraph{Impact on Semantic Segmentation} To further explore the benefits of SIM, we evaluate its impact on retinal vessel segmentation using four public datasets: DRIVE~\cite{drive}, STARE~\cite{stare}, CHASE-DB1~\cite{chasedb1}, and HRF~\cite{hrf}. Our baseline adopts a UNet architecture~\cite{ronneberger2015u} that is enhanced by three decoders, namely FCN~\cite{long2015fully}, DeepLabV3~\cite{chen2017rethinking}, and PSPNet~\cite{zhao2017pyramid}. 

As detailed in Table~\ref{table:seg}, integrating SIM consistently improves the Dice coefficient across most dataset-decoder configurations. For instance, UNet+FCN achieves 80.38\% on DRIVE, which is raised to 81.22\% with SIM, and similar gains are observed on STARE and CHASE-DB1. 

In summary, the empirical results demonstrate that SIM regularization enhances semantic retention by promoting coherent feature propagation. These consistent improvements across segmentation and translation tasks underscore SIM's potential as a general and effective mechanism for maintaining semantic fidelity in deep neural networks.

\paragraph{Ethical Considerations} While SIM promotes stronger semantic retention, the reconstruction modules could potentially allow partial recovery of sensitive inputs, raising privacy concerns in domains such as medical imaging, facial recognition, or speech data. To mitigate this risk, we recommend removing all reconstructors when releasing trained models.

\subsection{Can SIM Benefit Non-visual Domains?}

To assess SIM's generality beyond vision, we evaluate its impact on audio classification and multivariate time series anomaly detection, with detailed datasets and settings provided in the Appendix~\ref{sec:audio} and~\ref{sec:ts_ad} due to space limitations.

For audio classification on US8K~\cite{salamon2014dataset}, SCv2~\cite{warden2018speech}, and GTZAN~\cite{tzanetakis2002musical}, results in Table~\ref{tab:audio} in the Appendix show consistent gains in accuracy and F1 across backbones, with particularly pronounced improvements on the smaller GTZAN dataset, indicating enhanced temporal-spectral representations and mitigation of overfitting.

For time series anomaly detection on MSL~\cite{hundman2018detecting}, PSM~\cite{abdulaal2021practical}, SMAP~\cite{hundman2018detecting}, SMD~\cite{su2019robust}, and SWaT~\cite{mathur2016swat}, Table~\ref{tab:ts_ad} in the Appendix demonstrates consistent improvements, especially in F1-score, reflecting its ability to better preserve structural patterns in sequential data and capture subtle anomalies.

Together with vision tasks, these results indicate that SIM generalizes across modalities and effectively enhances semantic and structural feature retention in non-visual domains.

\subsection{Gradient Norm Analysis of SIM}
\label{sec:grad_norm}

This section investigates the impact of SIM on gradient norms during training. By enforcing feature reconstruction, SIM ensures that each layer retains critical input information, mitigating information loss and promoting gradient consistency across layers. 

Assume the existence of optimal mappings \( \mathcal{F} \) and \( \mathcal{G} \), such that \( \mathbf{x} = \mathcal{G}(\mathcal{F}(\mathbf{x})) \). For the composition \( \mathcal{G} \circ \mathcal{F} \), the Lipschitz constant satisfies:
\begin{equation}
	\| \mathcal{G}(\mathcal{F}(\mathbf{x_1})) - \mathcal{G}(\mathcal{F}(\mathbf{x_2})) \| \le L_{\mathcal{G}\circ\mathcal{F}}  \| \mathbf{x_1} - \mathbf{x_2} \|.
\end{equation}
with \( L_{\mathcal{G}\circ\mathcal{F}} = 1 \). Under SIM regularization, if the Lipschitz constant of \( \mathcal{F} \) is large, \( \mathcal{G} \) compensates by contracting the gradient flow, reducing distortions. If \( \mathcal{F} \)'s constant is small, \( \mathcal{G} \) accelerates the gradient flow, improving optimization speed. This interplay ensures stable and smooth gradient updates.

Large gradient norms have been associated with degraded convergence and poor generalization~\cite{shapiro1996convergence, zhao2022penalizing, xie2023overlooked}. \Cref{fig:grad_norm} illustrates the changes in gradient norms during training with the Swin Transformer on the CUB dataset. In the baseline, gradient norms exhibit frequent and sharp spikes, which which may destabilize optimization and hinder convergence. With SIM, these fluctuations are significantly reduced, resulting in smoother gradient flow and more consistent updates. This stabilization may contribute to the improved performance and generalization observed with SIM.

%% file: tables/exp_cifar_ext.tex
\begin{table*}[!h]
	\centering
	\renewcommand{\arraystretch}{1.1}
	\resizebox{0.98\linewidth}{!}{
		\begin{tabular}{ll||cccc|cccc|cc}
			\toprule
			\multirow{2}{*}{Backbone} & \multirow{2}{*}{Regularization} & \multicolumn{4}{c}{CIFAR-10} & \multicolumn{4}{c}{CIFAR-100} & \multicolumn{2}{c}{SVHN} \\
			\cline{3-6} \cline{7-10} \cline{11-12}
			& & Acc. & F1. & \#Params & $\Delta$ Acc. & Acc. & F1. & \#Params & $\Delta$ Acc. & Acc. & $\Delta$ Acc. \\
			\cline{1-12}
			\multirow{2}{*}{ResNet-18} 
			& Baseline & 94.65 & 94.65 & 11.17 & -- & 78.20 & 78.14 & 11.22 & -- & 96.29 & -- \\
			& $+\rho$SIM (Ours) & \textbf{94.94} & \textbf{94.94} & \textbf{11.23} & \textbf{0.29 $\uparrow$} & \textbf{78.68} & \textbf{78.62} & \textbf{11.28} & \textbf{0.48 $\uparrow$} & \textbf{96.41} & \textbf{0.12 $\uparrow$} \\
			\cline{1-12}
			\multirow{8}{*}{ResNet-50} 
			& Baseline & 95.26 & 95.26 & 23.52 & -- & 79.94 & 79.89 & 23.71 & -- & 96.76 & -- \\
			& $+\rho$SIM (Ours) & \textbf{95.39} & \textbf{95.38} & \textbf{24.23} & \textbf{0.13 $\uparrow$} & \textbf{80.03} & \textbf{79.97} & \textbf{24.42} & \textbf{0.09 $\uparrow$} & \textbf{96.87} & \textbf{0.11 $\uparrow$} \\
			& Mixup~\cite{zhang2017mixup} & 96.25 & 96.25 & 23.52 & -- & 81.16 & 81.12 & 23.71 & -- & 96.93 & -- \\
			& $+\rho$SIM (Ours) & \textbf{96.48} & \textbf{96.48} & \textbf{24.23} & \textbf{0.23 $\uparrow$} & \textbf{81.35} & \textbf{81.33} & \textbf{24.42} & \textbf{0.19 $\uparrow$} & \textbf{97.39} & \textbf{0.46 $\uparrow$} \\
			& CutMix~\cite{yun2019cutmix} & 96.60 & 96.60 & 23.52 & -- & 81.09 & 81.06 & 23.71 & -- & 97.68 & -- \\
			& $+\rho$SIM (Ours) & \textbf{96.69} & \textbf{96.69} & \textbf{24.23} & \textbf{0.09 $\uparrow$} & \textbf{81.66} & \textbf{81.67} & \textbf{24.42} & \textbf{0.57 $\uparrow$} & \textbf{97.74} & \textbf{0.06 $\uparrow$} \\
			& Label Smooth~\cite{szegedy2016rethinking} & 95.21 & 95.21 & 23.52 & -- & 79.99 & 80.01 & 23.71 & -- & 97.02 & -- \\
			& $+\rho$SIM (Ours) & \textbf{95.35} & \textbf{95.35} & \textbf{24.23} & \textbf{0.14 $\uparrow$} & \textbf{80.16} & \textbf{80.17} & \textbf{24.42} & \textbf{0.17 $\uparrow$} & \textbf{97.18} & \textbf{0.14 $\uparrow$} \\
			\cline{1-12}
			\multirow{2}{*}{ResNet-101} 
			& Baseline & 95.27 & 95.27 & 42.51 & -- & 79.97 & 79.92 & 42.70 & -- & 96.84 & -- \\
			& $+\rho$SIM (Ours) & \textbf{95.52} & \textbf{95.51} & \textbf{43.99} & \textbf{0.25 $\uparrow$} & \textbf{80.44} & \textbf{80.41} & \textbf{44.17} & \textbf{0.47 $\uparrow$} & \textbf{96.94} & \textbf{0.10 $\uparrow$} \\
			\cline{1-12}
			\multirow{2}{*}{ResNet-152} 
			& Baseline & 95.57 & 95.54 & 58.16 & -- & 80.19 & 80.14 & 58.34 & -- & 96.93 & -- \\
			& $+\rho$SIM (Ours) & \textbf{95.70} & \textbf{95.70} & \textbf{60.29} & \textbf{0.13 $\uparrow$} & \textbf{80.51} & \textbf{80.48} & \textbf{60.47} & \textbf{0.32 $\uparrow$} & \textbf{97.05} & \textbf{0.12 $\uparrow$} \\
			\bottomrule
		\end{tabular}
	}
	\caption{Image classification results on CIFAR-10 and CIFAR-100. We report the accuracy (\%), F1 score (\%), and number of parameters (in millions). $\Delta$ Acc. denotes the absolute improvement in accuracy (\%) relative to the corresponding baseline.}
	\label{tab:cifar}
\end{table*}

%% file: tables/exp_cub.tex
%
%
%
%
%

\begin{table}[!t]
	\centering
	\renewcommand{\arraystretch}{1.15}
	\resizebox{0.98\linewidth}{!}{
		\begin{tabular}{lcccc}
			\toprule
			Model & Acc. & \#Params & \#FLOPs & $\Delta$ Acc. \\
			\cline{1-5}
			ResNet-50~\cite{ResNet} & 87.99 & 23.92 & 12.14 & -- \\
			\bf $+\rho$SIM (Ours)  & \bf 88.25 & \bf 25.04 & \bf 12.46 & \bf 0.26 $\uparrow$ \\
			
			\cline{1-5}
			DenseNet-169~\cite{huang2017densely} & 84.52 & 12.82 & 10.11 & -- \\
			\bf $+\rho$SIM (Ours)  & \bf 84.67 & \bf 14.39 & \bf 10.26 & \bf 0.15 $\uparrow$ \\
			
			\cline{1-5}
			ConvNeXt~\cite{liu2022convnet} & 89.68 & 87.77 & 45.33 & -- \\
			\bf $+\rho$SIM (Ours)  & \bf 91.80 & \bf 89.56 & \bf 45.77 & \bf 2.12 $\uparrow$ \\
			
			\cline{1-5}
			Swin-B/16~\cite{liu2021swin} & 90.94 & 87.08 & 15.63 & -- \\
			\bf $+\rho$SIM (Ours)  & \bf 91.27 & \bf 87.51 & \bf 15.74 & \bf 0.33 $\uparrow$ \\
			
			\bottomrule
		\end{tabular}
	}
	\caption{Image classification results on the CUB dataset.}
	\label{tab:cub}
\end{table}

%% file: tables/exp_few_shot.tex
\begin{table*}[t]
	\centering
	\resizebox{0.76\textwidth}{!}{
		\begin{threeparttable}			
			\centering
			\setlength{\tabcolsep}{1.2mm}{
				\begin{tabular}{lcccccccc}
					\toprule
					Method                             & FGVC                    & OxfordPets & SUN397 & UCF101 & Caltech101 &  DTD   & EuroSAT  & Avg.  \\ \midrule
					Zero-shot CLIP                      & 0.42                     & 56.25      & 28.96  & 21.05  & 60.62      & 10.00   & 4.17     &  25.92 \\
					\midrule
					
					Tip-Adapter~\cite{zhang2021tip}                         & 77.19 &  89.64 &    66.25 &    92.86 &  34.86 &  84.65 &  71.36 &               73.83 \\ 
					
					\bf Tip-Adapter + Ours                         & \bf 77.48 &  \bf 89.81 &    \bf 66.67 &    \bf 93.02 &  \bf 36.21 &  \bf 84.88 &  \bf 71.43 &  \bf 74.21 \\
					\midrule 
					
					Meta-Adapter~\cite{song2023meta}                        &  47.58 &  \bf 86.49 &    49.58 &    70.00 &  19.38 &  68.75 &  51.20 &               56.14 \\
					
					\bf Meta-Adapter + Ours                        &  \bf 48.19 &  86.15 &    \bf 51.25 &    \bf 71.88 &  \bf 19.79 &  \bf 70.83 &  \bf 51.35 &  \bf 57.06 \\
					\bottomrule
			\end{tabular}}
	\end{threeparttable}}
	\caption{Comparison of cross-dataset generalization based on ImageNet~\cite{deng2009imagenet} pre-training in a 16-shot few-shot learning setting. The Tip-Adapter and Meta-Adapter are tuned on ImageNet and frozen for other datasets.}
	\label{tab:few_shot}
\end{table*}

%% file: tables/exp_dg_and_beta.tex
\begin{table*}[htbp]
	\begin{minipage}[!h]{0.38\textwidth}
		\centering
		\renewcommand{\arraystretch}{1.1}
		\resizebox{0.82\linewidth}{!}{
			\begin{tabular}[t]{lrr}
				\toprule
				Method &  PACS & TerraIncognita \\
				\midrule
				ERM & 85.37 & 46.89 \\				
				\bf $+\rho$SIM (Ours) & \bf 87.60 & \bf 47.98\\
				\cline{1-3}
				IRM~\cite{IRM} & 86.25 & 50.19 \\				
				\bf $+\rho$SIM (Ours) & \bf 87.44 & \bf 50.71\\
				\cline{1-3}
				MixStyle~\cite{mixstyle} & 86.02 & 46.81 \\				
				\bf $+\rho$SIM (Ours) & \bf 86.84 & \bf 47.09\\
				\cline{1-3}
				VREx~\cite{VREx} & 87.16 & 49.36 \\				
				\bf $+\rho$SIM (Ours) & \bf 87.30 & \bf 49.98\\
				\bottomrule
			\end{tabular}
		}
		\captionof{table}{Comparison of average accuracy (\%) across four domains in PACS and TerraIncognita with and without SIM for different DG algorithms (See \cref{tab:dg_full} in Appendix for full comparison).}
		\label{tab:dg}
	\end{minipage}
	\hspace{0.03\textwidth}
	\begin{minipage}[!h]{0.58\textwidth}
		\centering
		\includegraphics[width=\linewidth]{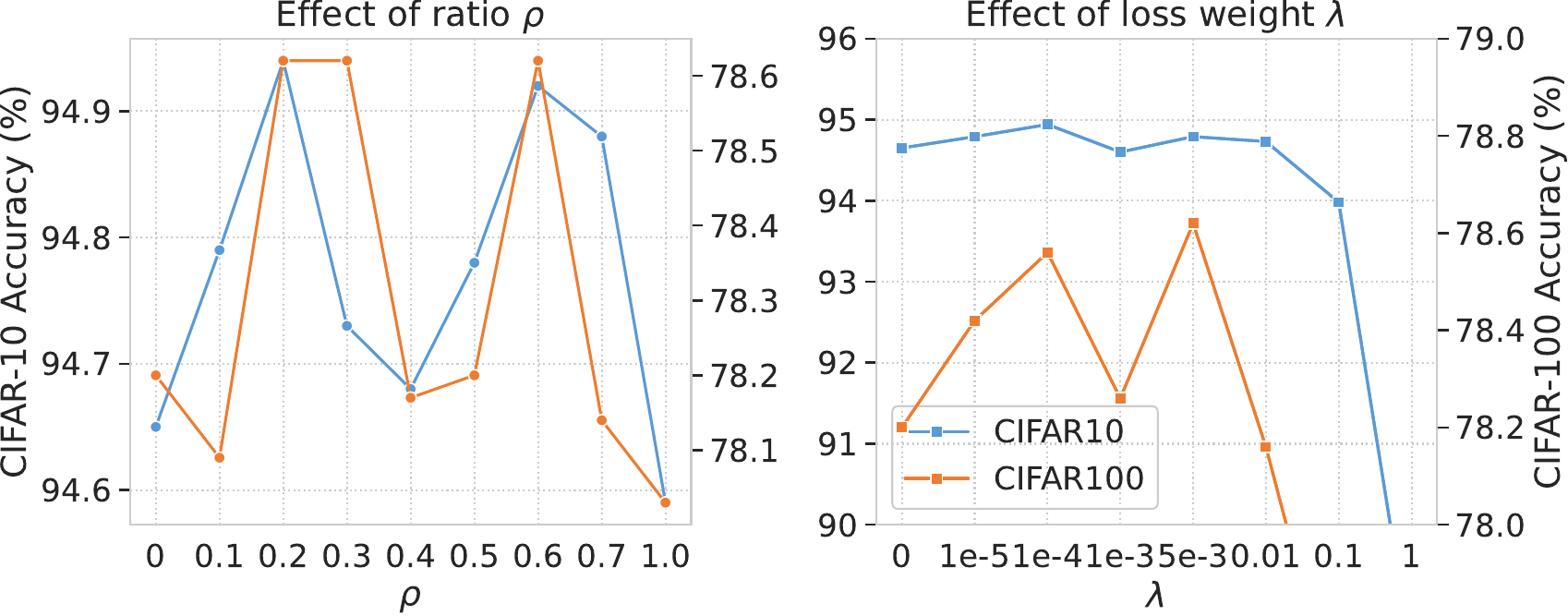}
		\captionof{figure}{Ablation study of token sampling ratio $\rho$ and SIM loss weight $\beta$ on ResNet18 for CIFAR-10 and CIFAR-100.}
		\label{fig:ablation}
	\end{minipage}
\end{table*}

%% file: tables/exp_i2i.tex
\begin{table*}[!t]
	\begin{minipage}[!h]{0.36\textwidth}
		\centering
		\renewcommand{\arraystretch}{1.1}
		\resizebox{0.99\linewidth}{!}{
			\begin{tabular}{{l}*{4}{c}}
				\toprule
				\multirow{2}*{Method} & \multicolumn{1}{c}{\textbf{Cat$\rightarrow$Dog}} &  \multicolumn{2}{c}{\textbf{Cityscapes}}\\
				\cmidrule(lr){2-2}\cmidrule(lr){3-4}
				& \textbf{FID} $\downarrow$ & \textbf{mAP} $\uparrow$ & \textbf{FID} $\downarrow$ \\
				
				\midrule
				CUT wo/ NCE & 126.5 & 4.6 & 364.8 \\
				CUT~\cite{park2020cut} & 76.2 & 24.7 & 56.4 \\ 
				\cline{1-4}
				$\rho$SIM + GAN Loss & \textbf{58.2} & \textbf{28.1} & \textbf{46.7} \\ 
				
				\bottomrule
			\end{tabular}
		}
		\captionof{table}{Quantitative comparison of different image translation methods. SIM modifies CUT~\cite{park2020cut} by simply removing the content constraint of PatchNCE and applying SIM regularization.}
		\label{tab:i2i}
	\end{minipage}
	\hspace{0.02\textwidth}
	\begin{minipage}[!h]{0.62\textwidth}
		\centering
		\includegraphics[width=0.96\linewidth]{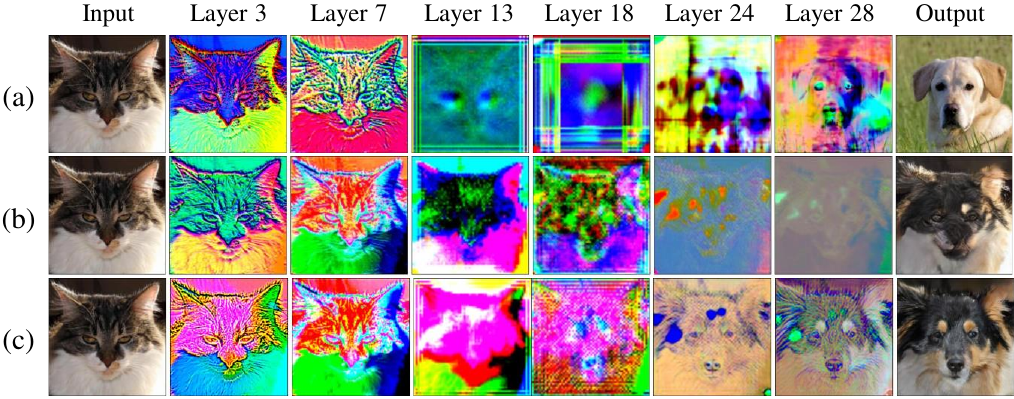}
		\captionof{figure}{PCA visualization of generator features at different layers. (a) Mode collapse, resulting from lack of content constraint. (b) CUT, using NCE loss to avoid mode collapse. (c) SIM, maintaining richer semantic structures and enhanced fine-grained details in the generated features.}
		\label{fig:pca}
	\end{minipage}
\end{table*}

%% file: tables/exp_drive.tex
\begin{table}[htbp]
	\centering	
	\renewcommand{\arraystretch}{1.1}
	\resizebox{0.98\linewidth}{!}{
		\begin{tabular}{lcccc}
			\toprule 
			Model &  DRIVE & STARE & CHASE-DB1 & HRF  \\
			\cline{1-5}
			UNet+FCN~\cite{long2015fully} &  80.38  & 81.22 & 80.31 & 80.45 \\
			\bf $+\rho$SIM (Ours)  & \bf 81.22 & \bf 81.32 & \bf 80.33  & \bf 80.52 \\
			
			\cline{1-5}
			UNet+DeepLabV3~\cite{chen2017rethinking} &  80.65  & 81.35 & \bf 80.82 & 80.34 \\
			\bf $+\rho$SIM (Ours)  & \bf 80.72 & \bf 81.46 & 80.45  & \bf 80.63 \\
			
			\cline{1-5}
			UNet+PSPNet~\cite{zhao2017pyramid} &  80.40  & 80.94 & 80.32 & 80.60 \\
			\bf $+\rho$SIM (Ours)  & \bf 80.64 & \bf 81.14 & \bf 80.36  & \bf 80.68 \\
			
			\bottomrule 
		\end{tabular}
	}
	\caption{Comparison of Dice scores (\%) for different segmentation networks across four public fundus segmentation datasets.}
	\label{table:seg}
\end{table}

%% file: sec/5_conclusion.tex
\section{Conclusion}
\label{sec:conclusion}

\begin{figure}[!t]
	\centering
	\includegraphics[width=0.98\linewidth]{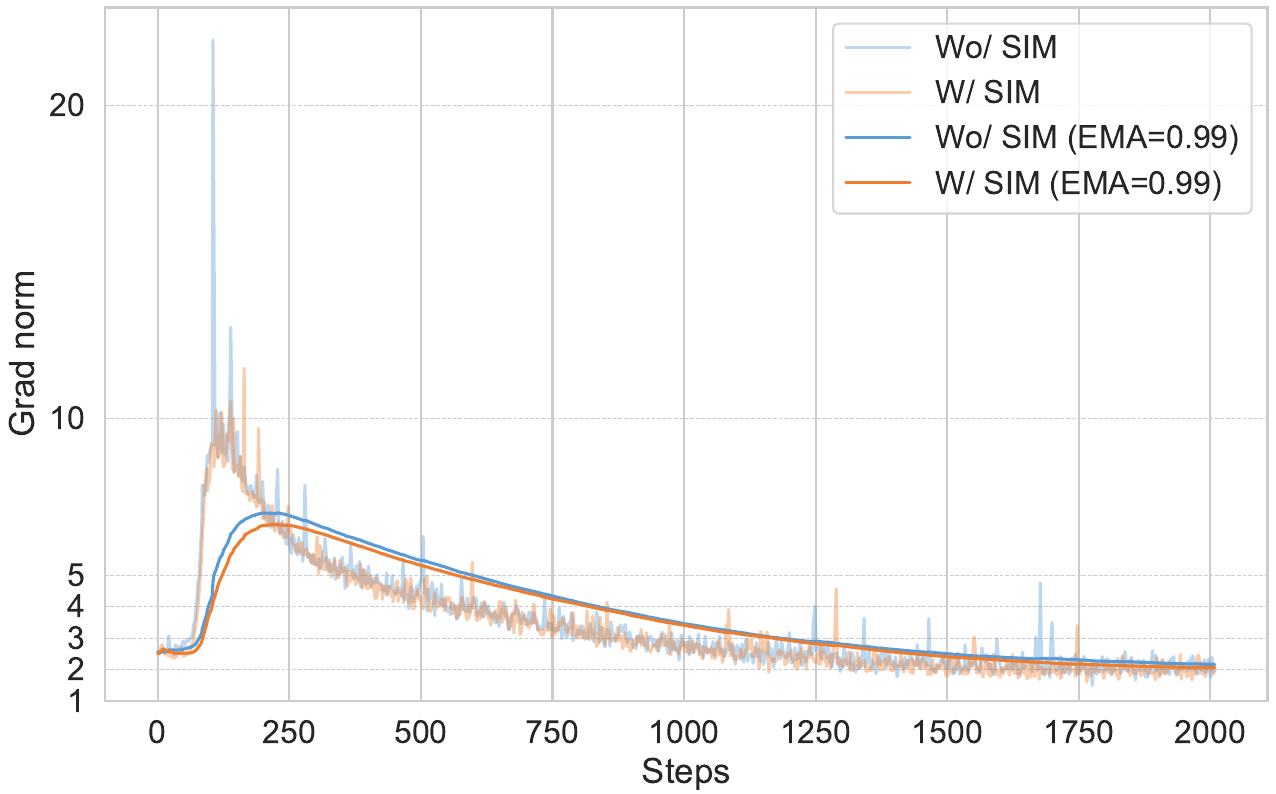}
	\caption{Gradient norm comparison during training of Swin-Transformer on CUB. EMA indicates exponential moving average.}
	\label{fig:grad_norm}
\end{figure}

In this paper, we introduce SIM, a simple yet effective data-intrinsic regularization framework that enhances feature learning via an inverse mapping mechanism. By reconstructing the input from its transformed output, SIM preserves progressively fine-grained features while mitigating information loss, thereby enhancing performance and improving generalization across diverse learning scenarios. To further improve efficiency, we propose $\rho\text{SIM}$, which incorporates token sampling and projection-based latent space reconstruction. These strategies significantly reduce computational complexity while maintaining strong performance, making SIM scalable and practical. We validate $\rho\text{SIM}$ across multiple tasks, including image classification, few-shot prompt tuning, and domain generalization, consistently demonstrating superior performance and strong generalization across various architectures. Moreover, our analysis shows that $\rho\text{SIM}$ effectively preserves semantic information, benefiting tasks requiring dense feature representations, and its applicability extends to non-visual domains, where it consistently improves feature representations across diverse datasets and architectures.

Despite these promising results, we acknowledge several limitations. First, due to computational constraints, our evaluation is mainly conducted on medium-scale datasets, which may not fully capture the behavior of SIM in large-scale scenarios. Future work will explore extending SIM to larger datasets and more complex domains, such as large language models and diffusion models. Second, while $\rho\text{SIM}$ alleviates the efficiency bottleneck, the reconstruction objective may prioritize certain features over others, leaving room for improved strategies that emphasize more distinctive and informative representations. We believe that addressing these aspects in future research will further enhance the applicability of SIM. We hope our work can stimulate further exploration in the regularization community.

%% file: sec/X_suppl.tex
\clearpage
\setcounter{page}{1}

\renewcommand{\thesubsection}{A.\arabic{subsection}} 

\section*{A. More Experiments}
\label{sec:appendix_A}

To further validate the generality of SIM beyond the visual domain, we conduct additional experiments on audio classification and multivariate time series anomaly detection tasks. These studies demonstrate that SIM consistently enhances representation learning across diverse modalities, confirming its broad applicability.

\subsection{Audio Classification}
\label{sec:audio}
We investigate the impact of SIM on audio classification using three benchmark datasets: US8K~\cite{salamon2014dataset}, Speech Commands v2 (SCv2)~\cite{warden2018speech}, and GTZAN~\cite{tzanetakis2002musical}. US8K is an environmental sound dataset with 10 classes and 8,732 audio clips, each lasting up to 4 seconds.
SCv2 is a large-scale keyword recognition dataset with 35 classes and over 105,000 utterances.
GTZAN is a relatively small dataset containing 1,000 music tracks evenly distributed across 10 genres. Audio classification presents unique challenges, as models must capture both temporal and spectral patterns from complex acoustic signals.

\paragraph{Baseline Methods} Four representative backbone architectures are considered: TDNN~\cite{martinez2022prediction}, CAM$++$~\cite{wang2023campp}, Res2Net~\cite{gao2019res2net}, and ERes2Net~\cite{chen2023enhanced}. Each model is evaluated in its original configuration, and $\rho$SIM is integrated into the backbone to assess its effect on feature consistency and discriminative capability.

\paragraph{Implementation Details} For each dataset, we extract log-mel filterbank (FBank) features as input representations. For SCv2, we use 40 filterbank bins, while for US8K and GTZAN, we use 80 bins. Models are trained with the same hyperparameters as their official implementations to ensure fair comparison. 

\paragraph{Results and Analysis} Table~\ref{tab:audio} reports accuracy and F1-score for all methods with and without $\rho$SIM. Overall, $\rho$SIM improves performance in 11 out of 12 architecture–dataset combinations, with only a slight drop for TDNN+SIM on GTZAN. Interestingly, the benefits of $\rho$SIM are particularly pronounced on GTZAN, which is a relatively small dataset containing only 1,000 music tracks. On this limited-scale benchmark, regularization plays a more critical role in preventing overfitting and enhancing generalization. For instance, CAM$++$ plus $\rho$SIM achieves absolute improvements of 1.78\% in accuracy and 5.86\% in F1, while ERes2Net+$\rho$SIM gains 3.00\% and 3.69\%, respectively. These results indicate that $\rho$SIM is particularly effective in data-scarce regimes, where reinforcing hierarchical feature representations enables models to capture richer temporal–spectral patterns from limited signals. This confirms that SIM generalizes effectively beyond visual tasks, acting as a versatile, task-agnostic regularization strategy applicable across different modalities.


\begin{table}[!t]
	\centering
	\renewcommand{\arraystretch}{1.}
	\begin{adjustbox}{max width=\linewidth}
		\begin{tabular}{l cc cc cc}
			\toprule[1pt]
			\multirow{2}{*}{\textbf{Algorithm}} & \multicolumn{2}{c}{\textbf{US8K}} & \multicolumn{2}{c}{\textbf{SCv2}} & \multicolumn{2}{c}{\textbf{GTZAN}} \\
			\cmidrule(lr){2-3} \cmidrule(lr){4-5} \cmidrule(lr){6-7}
			& \textbf{Acc.} & \textbf{F1.} & \textbf{Acc.} & \textbf{F1.} & \textbf{Acc.} & \textbf{F1.} \\
			\midrule
			TDNN~\cite{martinez2022prediction} & 94.89 & 94.95 & 97.94 & 96.60 & 71.43 & \textbf{71.68} \\
			\bf $+\rho$SIM (Ours) & \textbf{95.02} & \textbf{95.13} & \textbf{97.98} & \textbf{96.72} & \textbf{73.21} & 70.59 \\
			$\Delta$ & 
			\textcolor{red}{+0.13} & \textcolor{red}{+0.18} & \textcolor{red}{+0.04} & \textcolor{red}{+0.12} & \textcolor{red}{+1.78} & \textcolor{blue}{-1.09} \\
			\midrule
			CAM$++$~\cite{wang2023campp} & 96.07 & 95.97 & 97.46 & 95.85 & 74.11 & 70.67 \\
			\bf $+\rho$SIM (Ours) & \textbf{96.34} & \textbf{96.47} & \textbf{97.58} & \textbf{96.07} & \textbf{75.89} & \textbf{76.53} \\
			$\Delta$ & 
			\textcolor{red}{+0.27} & \textcolor{red}{+0.50} & \textcolor{red}{+0.12} & \textcolor{red}{+0.22} & \textcolor{red}{+1.78} & \textcolor{red}{+5.86} \\
			\midrule
			Res2Net~\cite{gao2019res2net} & 92.74 & 92.90 & 96.51 & 94.01 & 69.64 & 69.54 \\
			\bf $+\rho$SIM (Ours) & \textbf{93.99} & \textbf{94.03} & \textbf{96.72} & \textbf{94.73} & \textbf{70.54} & \textbf{72.11} \\
			$\Delta$ & 
			\textcolor{red}{+1.25} & \textcolor{red}{+1.13} & \textcolor{red}{+0.21} & \textcolor{red}{+0.72} & \textcolor{red}{+0.90} & \textcolor{red}{+2.57} \\
			\midrule
			ERes2Net~\cite{chen2023enhanced} & 96.29 & 96.57 & 98.22 & 97.11 & 74.00 & 71.81 \\
			\bf $+\rho$SIM (Ours) & \textbf{96.81} & \textbf{97.12} & \textbf{98.43} & \textbf{97.47} & \textbf{77.00} & \textbf{75.50} \\
			$\Delta$ & 
			\textcolor{red}{+0.52} & \textcolor{red}{+0.55} & \textcolor{red}{+0.21} & \textcolor{red}{+0.36} & \textcolor{red}{+3.00} & \textcolor{red}{+3.69} \\
			\toprule[1pt]
		\end{tabular}
	\end{adjustbox}
	\caption{Comparison of accuracy (\%) and F1-score (\%) on US8K~\cite{salamon2014dataset}, SCv2~\cite{warden2018speech}, and GTZAN~\cite{tzanetakis2002musical} datasets for audio classification task.}
	\label{tab:audio}
\end{table}

\begin{table*}[!t]	
	\centering
	\begin{adjustbox}{max width=\linewidth}
		\begin{tabular}{lcccccccccc}
			\toprule
			\multirow{2}{*}{\textbf{Models}} 
			& \multicolumn{2}{c}{\textbf{MSL}} 
			& \multicolumn{2}{c}{\textbf{PSM}} 
			& \multicolumn{2}{c}{\textbf{SMAP}} 
			& \multicolumn{2}{c}{\textbf{SMD}} 
			& \multicolumn{2}{c}{\textbf{SWAT}} \\
			\cmidrule(lr){2-3} \cmidrule(lr){4-5} \cmidrule(lr){6-7} \cmidrule(lr){8-9} \cmidrule(lr){10-11}
			& \textbf{Acc.} & \textbf{F1.} & \textbf{Acc.} & \textbf{F1.} & \textbf{Acc.} & \textbf{F1.} & \textbf{Acc.} & \textbf{F1.} & \textbf{Acc.} & \textbf{F1.} \\
			\midrule
			
			MambaTS~\cite{cai2024mambats} & 95.86 & 78.13 & 97.93 & 96.20 & \textbf{93.55} & 68.64 & 98.75 & 84.48 & \textbf{98.57} & 94.22 \\
			\bf $+\rho$SIM (Ours) & \textbf{96.45} & \textbf{81.68} & \textbf{97.96} & \textbf{96.24} & \textbf{93.55} & \textbf{68.80} & \textbf{98.77} & \textbf{84.74} & \textbf{98.57} & \textbf{94.23} \\
			$\Delta$ & \textcolor{red}{+0.59} & \textcolor{red}{+3.55} & \textcolor{red}{+0.03} & \textcolor{red}{+0.04} & 0.00 & \textcolor{red}{+0.16} & \textcolor{red}{+0.02} & \textcolor{red}{+0.26} & 0.00 & \textcolor{red}{+0.01} \\
			\midrule
			iTransformer~\cite{liu2023itransformer} & 94.98 & 72.38 & 97.27 & 94.92 & 93.27 & 66.73 & 98.50 & 80.84 & \textbf{98.22} & \textbf{92.69} \\
			\bf $+\rho$SIM (Ours) & \textbf{96.06} & \textbf{79.50} & \textbf{97.77} & \textbf{95.88} & \textbf{93.35} & \textbf{67.30} & \textbf{98.71} & \textbf{83.77} & 98.21 & 92.67 \\
			$\Delta$ & \textcolor{red}{+1.08} & \textcolor{red}{+7.12} & \textcolor{red}{+0.50} & \textcolor{red}{+0.96} & \textcolor{red}{+0.08} & \textcolor{red}{+0.57} & \textcolor{red}{+0.21} & \textcolor{red}{+2.93} & \textcolor{blue}{-0.01} & \textcolor{blue}{-0.02} \\
			\midrule
			LightTS~\cite{zhang2207less} & 96.31 & 80.82 & \textbf{97.79} & \textbf{95.92} & \textbf{93.35} & 67.49 & 98.62 & 82.59 & 98.22 & 92.69 \\
			\bf $+\rho$SIM (Ours) & \textbf{96.39} & \textbf{81.29} & 97.61 & 95.56 & \textbf{93.35} & \textbf{67.51} & \textbf{98.68} & \textbf{83.48} & \textbf{98.85} & \textbf{95.29} \\
			$\Delta$ & \textcolor{red}{+0.08} & \textcolor{red}{+0.47} & \textcolor{blue}{-0.18} & \textcolor{blue}{-0.36} & 0.00 & \textcolor{red}{+0.02} & \textcolor{red}{+0.06} & \textcolor{red}{+0.89} & \textcolor{red}{+0.63} & \textcolor{red}{+2.60} \\
			\midrule
			TimesNet~\cite{wu2022timesnet} & 96.22 & 80.47 & 96.70 & 93.77 & 93.57 & 68.90 & \textbf{98.61} & 82.24 & 98.19 & 92.59 \\
			\bf $+\rho$SIM (Ours) & \textbf{96.23} & \textbf{80.48} & \textbf{97.52} & \textbf{95.38} & \textbf{93.61} & \textbf{69.22} & \textbf{98.61} & \textbf{82.31} & \textbf{98.21} & \textbf{92.68} \\
			$\Delta$ & \textcolor{red}{+0.01} & \textcolor{red}{+0.01} & \textcolor{red}{+0.82} & \textcolor{red}{+1.61} & \textcolor{red}{+0.04} & \textcolor{red}{+0.32} & 0.00 & \textcolor{red}{+0.07} & \textcolor{red}{+0.02} & \textcolor{red}{+0.09} \\
			\bottomrule
		\end{tabular}
	\end{adjustbox}
	\caption{
		Multivariate time series anomaly detection results (Accuracy and F1-score). on five benchmark datasets: MSL~\cite{hundman2018detecting}, PSM~\cite{abdulaal2021practical}, SMAP~\cite{hundman2018detecting}, SMD~\cite{su2019robust}, and SWAT~\cite{mathur2016swat}.
		\(\Delta\) indicates the performance change after adding SIM: red for improvement, blue for degradation, and black for no change.
	}
	\label{tab:ts_ad}
\end{table*}

\subsection{Time Series Anomaly Detection}
\label{sec:ts_ad}
To assess the generality of SIM on sequential data, we evaluate its performance on five widely used multivariate time series anomaly detection benchmarks. The MSL dataset~\cite{hundman2018detecting}, collected from NASA's Mars Science Laboratory, comprises 55 telemetry channels with 58,317 normal training samples and 73,729 testing samples, yielding an anomaly ratio of 10.72\%. The SMAP dataset~\cite{hundman2018detecting}, derived from the Soil Moisture Active Passive satellite, contains 25 variables with 135,183 training and 427,617 testing samples, with 12.13\% anomalies. The SMD dataset~\cite{su2019robust}, obtained from a large-scale IT system, includes 38 dimensions with 708,405 training and 708,420 testing samples, and an anomaly rate of 4.16\%. The PSM dataset~\cite{abdulaal2021practical}, sourced from a power system monitoring application, consists of 26 variables with 132,481 training and 87,841 testing samples. The SWaT dataset~\cite{mathur2016swat}, collected from a secure water treatment testbed, records 51 sensors and comprises 495,000 normal and 449,919 anomalous samples, where anomalies arise from both cyber-attacks and natural faults. Accurate detection in these tasks requires capturing complex temporal dependencies while remaining robust to noise and missing values. By including these datasets, we assess SIM's ability to generalize across different domains and anomaly types in multivariate sequential data.

\paragraph{Baseline Methods} We select four representative models: MambaTS~\cite{cai2024mambats}, iTransformer~\cite{liu2023itransformer}, LightTS~\cite{zhang2207less}, and TimesNet~\cite{wu2022timesnet}. Each baseline is evaluated in its standard configuration, and we integrate $\rho$SIM into the backbone to assess its contribution to hierarchical feature consistency and anomaly discriminability.

\paragraph{Implementation Details} We follow the open-source repository \texttt{Time-Series-Library}\footnote{\url{https://github.com/thuml/Time-Series-Library}}
to ensure consistency with prior works. Following the widely adopted Anomaly-Transformer paradigm~\cite{xu2022anomaly}, models are trained only on normal samples and evaluated by distinguishing anomalous patterns at test time. For fair comparison, all models are trained with the official hyperparameters of their original implementations. $\rho$SIM is incorporated as an auxiliary regularization module during training, reconstructing intermediate temporal features to encourage progressively refined representations and no modifications are made to the inference pipeline.

\paragraph{Comparison Results} Table~\ref{tab:ts_ad} reports accuracy and F1-score for all models with and without $\rho$SIM. Overall, $\rho$SIM consistently improves performance across most datasets and architectures. For instance, MambaTS gains +0.59\% Acc. and +3.55\% F1 on MSL, and iTransformer achieves +1.08\% Acc. and +7.12\% F1 on the same dataset. Improvements are particularly notable in F1-score, highlighting more accurate detection of anomalous sequences. Although a slight decrease is observed on SWAT with iTransformer, the overall trend demonstrates that $\rho$SIM enhances temporal feature representations and helps models capture subtle deviations and structural patterns critical for robust anomaly detection. These results further reinforce the versatility and modality-agnostic nature of SIM as a general regularization strategy.

\renewcommand{\thesubsection}{B.\arabic{subsection}} 

\section*{B. More Ablations}
\label{sec:appendix_B}

\subsection{Loss Function for SIM}
\label{sec:loss_type}

In order to investigate the effect of different loss functions in our proposed SIM regularization, we conduct an ablation study on CIFAR-10 and CIFAR-100 using ResNet18. The baseline model refers to the standard training setup without incorporating SIM. We evaluate several commonly used loss functions, including Cosine Similarity, KL Divergence, Wasserstein Distance, $\ell_1$ Loss, and Mean Squared Error (MSE) Loss, in the context of our regularization framework. As shown in Table \ref{tab:sim_ablation}, the baseline achieves 94.65\% accuracy on CIFAR-10. Among the tested loss functions, KL Divergence (94.85\%) and $\ell_1$ Loss (94.77\%) show marginal improvements, whereas Wasserstein Distance (94.50\%) and Cosine Similarity (94.54\%) do not contribute positively. Standard MSE Loss achieves 94.66\%, which is comparable to the baseline but does not show a significant improvement.

Notably, when applying normalized MSE Loss, the accuracy improves to 94.94\%, achieving the best performance among all configurations. This suggests that normalization plays a crucial role in stabilizing the optimization process and enhancing feature consistency, thereby improving generalization.

\input{tables/exp_ablation}

\subsection{Efficiency Analysis}
\label{sec:eff}

\begin{table*}[htbp]
	\centering	
	\begin{tabular}{lccc ccc}
		\toprule
		\multirow{2}{*}{Model} & \multicolumn{3}{c}{Training Speed (iter/sec)} & \multicolumn{3}{c}{GPU Memory (MB)} \\
		\cmidrule(lr){2-4} \cmidrule(lr){5-7}
		& Baseline & $+\rho$SIM & Increase Ratio & Baseline & $+\rho$SIM & Increase Ratio \\
		\midrule
		ResNet-18     & 0.0864 & 0.0961 & 1.11 &  2901 &  3025 & 1.04 \\
		ResNet-34     & 0.1456 & 0.1650 & 1.13 &  4250 &  4477 & 1.05 \\
		ResNet-50     & 0.2459 & 0.2890 & 1.18 & 10876 & 11615 & 1.07 \\
		ResNet-101    & 0.3840 & 0.4622 & 1.20 & 16100 & 17263 & 1.07 \\
		ConvNeXt & 0.6362 & 0.6985 & 1.10 & 22261 & 22471 & 1.01 \\
		DenseNet-169 & 0.3337 & 0.3485 & 1.04 & 19462 & 19707 & 1.01 \\
		Swin-B/16   & 0.4099 & 0.4276 & 1.04 & 13372 & 13535 & 1.01 \\
		Swin-L/32   & 0.7462 & 0.7822 & 1.05 & 21077 & 21297 & 1.01 \\
		\bottomrule
	\end{tabular}
	\caption{Experimental comparison between Baseline and SIM with respect to training time and GPU memory usage.}
	\label{tab:efficiency}
\end{table*}

\input{tables/dg_full}

In Table~\ref{tab:efficiency}, we further analyze the impact of SIM (with the default parameter $\rho = 0.2$) on training speed and GPU memory consumption on the CUB-200 dataset. Overall, SIM introduces additional training overhead, and this effect becomes more pronounced as the network depth and size increase. This observation is consistent with our expectation: since SIM requires an additional reconstruction branch during training, it theoretically doubles the computational and memory cost. However, thanks to the sampling strategy of $\rho$-SIM and its latent-space reconstruction design, the actual overhead is significantly lower than the theoretical bound. For instance, on ResNet-101, the training speed decreases by only about 20\%, while for smaller models such as ResNet-18 and ResNet-34, the additional training time is usually limited to around 10\%. On the other hand, compared to training speed, the increase in GPU memory usage is much smaller, typically ranging from 1\% to 7\%. Interestingly, we also observe that Swin Transformers incur lower overhead than convolutional networks when applying SIM. This is mainly due to the sampling strategy: for convolutional networks, dense feature maps result in more sampled elements under the same ratio, whereas for Swin Transformers, inputs are first divided into patches, and with $\rho = 0.2$, the number of sampled patches is much smaller than the number of sampled pixels in CNN feature maps, leading to lower overall cost. These findings indicate that SIM improves model expressiveness while keeping its additional training cost practically manageable.

\section*{C. Full Experimental Results}
\label{sec:appendix_C}
We present the complete domain generalization experimental results in Table~\ref{tab:dg_full}.

\section*{D. Datasets}
\label{sec:appendix_D}

We provide detailed dataset splits and the number of categories in Table~\ref{tab:datasets}.

\renewcommand{\thesubsection}{E.\arabic{subsection}} 

\section*{E. Additional Experimental Details}
\label{sec:appendix_E}

\input{tables/datasets}

\subsection{Image Classification}
\label{sec:appendix_ic}


For the image classification experiments, we utilize the mmpretrain library\footnote{\url{https://github.com/open-mmlab/mmpretrain}}, which provides a comprehensive framework for training and evaluating deep learning models on image classification tasks. 

The experiments are conducted on three datasets: CIFAR-10, CIFAR-100, SVHN, and CUB. For CIFAR-10 and SVHN, the model is trained using the stochastic gradient descent (SGD) optimizer with a learning rate of 0.1, momentum of 0.9, and weight decay of 0.0001. The learning rate follows a multi-step decay, reducing by a factor of 0.1 at the 100th and 150th epochs, with a gamma of 0.1. Training is conducted for 200 epochs with a batch size of 128. For CIFAR-100, the configuration is adjusted to account for the increased complexity of the dataset. The optimizer uses a higher weight decay of 0.0005, and the learning rate scheduler reduces the learning rate at epochs 60, 120, and 160, with a gamma of 0.2. The batch size remains 128, consistent with CIFAR-10.

For the fine-grained classification task on the CUB dataset, we use ResNet50 with pre-trained weights converted from ImageNet21K\footnote{\url{https://github.com/Alibaba-MIIL/ImageNet21K}}. The training configuration is as follows: the optimizer is set to SGD with a learning rate of 0.01, momentum of 0.9, weight decay of 0.0005, and Nesterov acceleration. The learning rate follows a warm-up strategy for the first 5 epochs, gradually increasing from 0.01 to the base learning rate. Afterward, a cosine annealing scheduler is used for the remaining epochs, with the learning rate gradually decaying over 95 epochs. The model is trained for 100 epochs with a batch size of 64. For the Swin-Transformer, we use pre-trained weights provided by the official repository\footnote{\url{https://github.com/microsoft/Swin-Transformer}}. The optimizer is set to AdamW with a learning rate of 5e-6, weight decay of 0.0005, and a clip gradient value of 5.0. The learning rate is scheduled using the AdamW optimizer with custom settings, including specific parameter-wise learning rate adjustments for certain layers. Additionally, the training configuration includes logging every 20 intervals and saving the last three checkpoints. The model is trained for a maximum of 100 epochs with a batch size of 64.

\subsection{Few-Shot Image Classification}
\label{sec:appendix_fs}


To ensure a fair comparison, we adopt two baseline methods for few-shot image classification: Tip-Adapter\footnote{\url{https://github.com/gaopengcuhk/Tip-Adapter}} and Meta-Adapter\footnote{\url{https://github.com/ArsenalCheng/Meta-Adapter}}. Both methods are renowned for their ability to perform few-shot learning without requiring additional training. For Tip-Adapter, we used the official configuration provided by the authors, which employs a training-free adapter to adapt vision-language models. Similarly, for Meta-Adapter, we followed the official guidelines to ensure consistency with the original experiments. These baselines serve as a reference to evaluate the performance of our method under consistent experimental setup.

\subsection{Domain Generalization}
\label{sec:appendix_dg}


For the domain generalization (DG) experiments, we primarily use the Transfer Learning Library\footnote{\url{https://github.com/thuml/Transfer-Learning-Library}}, conducting each experiment with three random seeds and reporting the average results. The same experimental setup is used for both DG and DG+SIM comparisons.

For the ERM algorithm, the training configuration includes an initial learning rate of 1e-3, momentum of 0.9, weight decay of 0.0005, and a batch size of 36. The training is conducted for 20 epochs, with 500 iterations per epoch. These parameters are consistent with the standard configuration for domain generalization tasks, ensuring comparability with previous methods.

\subsection{Image Segmentation}
\label{sec:appendix_is}


For the fundus segmentation experiment, we primarily use the mmsegmentation framework\footnote{\url{https://github.com/open-mmlab/mmsegmentation}}. This library offers a flexible and efficient platform for semantic segmentation tasks, supporting various advanced models and datasets.

\subsection{Image-to-Image Translation}
\label{sec:appendix_i2}


For the image translation task, the CUT model is implemented using the official repository\footnote{\url{https://github.com/taesungp/contrastive-unpaired-translation}}. The Fréchet Inception Distance (FID) is computed using the pytorch-fid library\footnote{\url{https://github.com/mseitzer/pytorch-fid}}, which is widely used for evaluating the quality of generated images. These libraries are chosen due to their robust and reliable performance in image translation tasks, providing accurate and standardized evaluations.

%% file: tables/exp_ablation.tex
\begin{table}[!h]
	\centering
	\renewcommand{\arraystretch}{1.1}
	\resizebox{0.86\linewidth}{!}{ 
		\begin{tabular}{lcc}
			\toprule
			Loss Type & CIFAR-10 & CIFAR-100 \\
			\midrule
			Baseline (Wo/ SIM) & 94.65 & 78.20 \\
			\midrule
			Cosine Similarity & 94.54 & 78.36 \\
			KL Divergence & 94.85 & 76.89 \\
			Wasserstein Distance & 94.50 & 78.03 \\
			$\ell_1$ Loss & 94.77 & 78.32 \\
			MSE Loss & 94.66 & 78.38 \\
			\midrule
			\textbf{MSE Loss (Normalized)} & \textbf{94.94} & \textbf{78.68} \\
			\bottomrule
		\end{tabular}
	}
	\caption{Ablation study of different loss functions in SIM on ResNet18. We report classification accuracy (\%) on CIFAR-10 and CIFAR-100 datasets. The best result is highlighted in \textbf{bold}.}
	\label{tab:sim_ablation}
\end{table}

%% file: tables/dg_full.tex
\begin{table*}[!ht]
	\small
	\centering
	\renewcommand{\arraystretch}{1.}
	\resizebox{0.8\textwidth}{!}{
		\begin{tabular}{l ccccc ccccc}
			\toprule[1pt]
			\multirow{2}{*}{\textbf{Algorithm}} & \multicolumn{5}{c}{\textbf{PACS}} & \multicolumn{5}{c}{\textbf{TerraIncognita}} \\
			\cmidrule(lr){2-6} \cmidrule(lr){7-11}
			& \textbf{P} & \textbf{A} & \textbf{C} & \textbf{S} & \textbf{AVG} & \textbf{L100} & \textbf{L38} & \textbf{L43} & \textbf{L46} & \textbf{AVG} \\
			\midrule
			ERM & 97.01 & 87.74 & 79.61 & 77.12 & 85.37 & 41.16 & 55.65 & 38.97 & \textbf{51.79} & 46.89 \\
			\bf $+\rho$SIM (Ours) & \textbf{97.54} & \textbf{88.87} & \textbf{82.30} & \textbf{81.70} & \textbf{87.60} & \textbf{44.27} & \textbf{57.36} & \textbf{40.51} & 49.78 & \textbf{47.98} \\
			\midrule
			IRM & 97.49 & 88.96 & \textbf{82.50} & 76.05 & 86.25 & 46.31 & \textbf{59.75} & \textbf{44.59} & 50.09 & 50.19 \\
			\bf $+\rho$SIM (Ours) & \textbf{97.96} & \textbf{89.79} & 81.57 & \textbf{80.43} & \textbf{87.44} & \textbf{48.32} & 59.63 & 42.06 & \textbf{52.82} & \textbf{50.71} \\
			\midrule
			MixStyle & 94.97 & \textbf{87.06} & 79.74 & 82.31 & 86.02 & \textbf{43.36} & \textbf{56.32} & 32.60 & \textbf{54.97} & 46.81 \\
			\bf $+\rho$SIM (Ours) & \textbf{95.75} & 86.91 & \textbf{81.40} & \textbf{83.30} & \textbf{86.84} & 42.99 & 54.35 & \textbf{36.56} & 54.46 & \textbf{47.09} \\
			\midrule
			VREx & 97.37 & 87.16 & 81.48 & \textbf{82.62} & 87.16 & \textbf{46.27} & 58.46 & 40.95 & 51.77 & 49.36 \\
			\bf $+\rho$SIM (Ours) & \textbf{97.54} & \textbf{89.06} & \textbf{81.57} & 81.04 & \textbf{87.30} & 42.90 & \textbf{58.73} & \textbf{44.12} & \textbf{54.17} & \textbf{49.98} \\
			\toprule[1pt]
		\end{tabular}
	}
	\caption{Full comparison of accuracy (\%) across domains in PACS and TerraIncognita with and without SIM for different DG algorithms.}
	\label{tab:dg_full}
\end{table*}

%% file: tables/datasets.tex
\begin{table*}[ht] 
	\small
	\centering
	\resizebox{0.88\textwidth}{!}{
		\begin{tabular}{l r r r r r}
			\toprule
			Dataset & Classes & Train examples & Valid.\,examples & Test examples & Accuracy measure\\
			\midrule			
			CIFAR-10~\cite{cifar}  &$10$ & $45000$ & $5000$ &$10000$ & Top-1 Acc. / F1 score \\
			CIFAR-100~\cite{cifar} &$100$ & $44933$ & $5067$ &$10000$ & Top-1 Acc. / F1 score \\		
			SVHN~\cite{svhn}  &$10$ & $73257$ & $26032$ &$26032$ & Top-1 Acc. \\
			CUB-200~\cite{cub200} & $200$ & $5994$ & $5794$ & $5794$ & Top-1 Acc. \\	
			Sun397~\cite{sun397}  &$397$ & $15880$ &$3970$ &$19850$ & Top-1 Acc. \\
			FGVCAircraft~\cite{aircraft} & $100$ & $3334$ & $3333$ & $3333$ & Top-1 Acc.  \\			
			DTD ~\cite{dtd} & $47$ & $1880$ & $1880$ & $1880$ & Top-1 Acc. \\
			OxfordPets~\cite{pets} & $37$ & $2940$ & $740$ & $3669$ & Top-1 Acc. \\
			Caltech-101~\cite{caltech101} & $101$ & $2550$ & $510$ & $6084$ & Top-1 Acc. \\
			UCF101~\cite{soomro2012ucf101} & $101$ & $9537$ & $3783$ & $3783$ & Top-1 Acc. \\
			EuroSAT~\cite{helber2019eurosat} & $10$ & $13500$ & $5400$ & $8100$ & Top-1 Acc. \\
			Dog-to-Cat~\cite{choi2020stargan} & -- & $9892$ & $500$ & $500$ & FID. \\
			CityScapes~\cite{cordts2016cityscapes} & 21 & $2975$ & $500$ & $500$ & FID. / mAP \\
			VOC-2007 ~\cite{pascal} &$20$ & $2501$ &$2510$ &$4952$ & CF1 / OF1 / mAP \\
			
			
			\bottomrule
		\end{tabular}
	}
	\vspace{0.5em}
	\captionsetup{width=.9\linewidth}
	\caption{Detailed dataset splits and the number of categories.}
	\label{tab:datasets}
\end{table*}